\documentclass[runningheads]{llncs}

\usepackage{eccvabbrv}

\usepackage{graphicx}
\usepackage{booktabs}
\usepackage{multirow}

\usepackage[accsupp]{axessibility}

\usepackage[pagebackref,breaklinks,colorlinks]{hyperref}
\usepackage{cleveref}

\usepackage{orcidlink}

\begin{document}

\title{Pose-Guided Fine-Grained Sign Language Video Generation} 

\titlerunning{Pose-Guided Sign Language Video Generation} 

\author{Tongkai Shi\inst{1} \and
Lianyu Hu\inst{1} \and
Fanhua Shang\inst{1} \and
Jichao Feng\inst{1} \and
Peidong Liu\inst{1} \and
Wei Feng\inst{1}}

\authorrunning{T. Shi et al.}

\institute{Tianjin University, Tianjin, China\\
\email{\{stk,hly2021,fhshang,jcfeng,peidongliu\_2023\}\@tju.edu.cn} \\
\email{wfeng@ieee.org}}

\maketitle

\begin{abstract}
Sign language videos are an important medium for spreading and learning sign language. However, most existing human image synthesis methods produce sign language images with details that are distorted, blurred, or structurally incorrect. They also produce sign language video frames with poor temporal consistency, with anomalies such as flickering and abrupt detail changes between the previous and next frames. To address these limitations, we propose a novel Pose-Guided Motion Model (PGMM) for generating fine-grained and motion-consistent sign language videos. Firstly, we propose a new Coarse Motion Module (CMM), which completes the deformation of features by optical flow warping, thus transfering the motion of coarse-grained structures without changing the appearance; Secondly, we propose a new Pose Fusion Module (PFM), which guides the modal fusion of RGB and pose features, thus completing the fine-grained generation. Finally, we design a new metric, Temporal Consistency Difference (TCD) to quantitatively assess the degree of temporal consistency of a video by comparing the difference between the frames of the reconstructed video and the previous and next frames of the target video. Extensive qualitative and quantitative experiments show that our method outperforms state-of-the-art methods in most benchmark tests, with visible improvements in details and temporal consistency.
  \keywords{Sign Language Video Generation \and Fine-Grained \and Temporal Consistency \and Modal Fusion}
\end{abstract}

\section{Introduction}
\label{sec:intro}

Sign language is the primary means of communication for the deaf community, characterized by rich and intricate visual information. Utilizing deep learning methods to synthesize artificial sign language videos from a reference image and motion data, a task known as Sign Language Video Generation (SLVG)\cite{suo2022jointly}, is crucial for deaf individuals to integrate into predominantly spoken language environments. Given that sign language semantics rely on detailed hand movements, the coherence and precision of hand details become essential requirements for SLVG.

\begin{figure}[tb]
  \centering
  \includegraphics[width=1\textwidth]{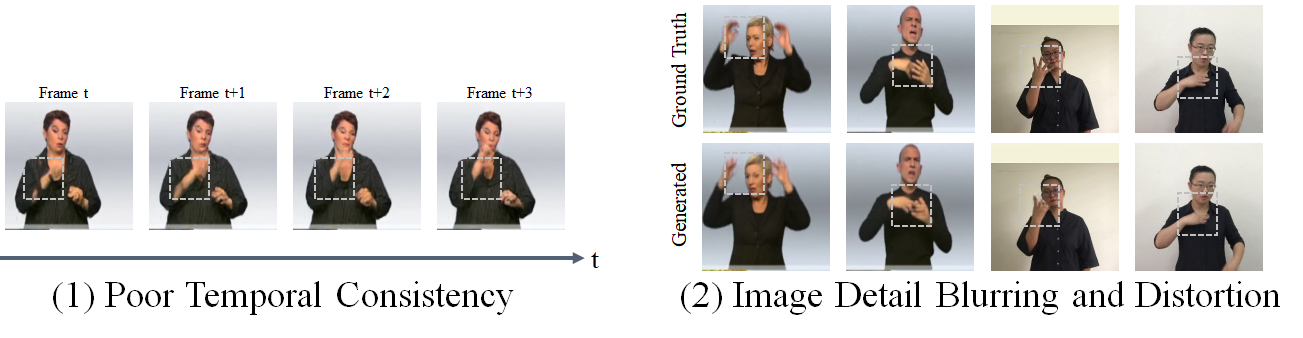}
  \caption{
    Problems of existing methods: (1) Poor temporal consistency. (2) Image detail blurring and distortion
  }
  \label{fig:problem}
\end{figure}

Most of the existing sign language video generation methods rely on human image synthesis\cite{jia2022human} methods, however, these methods are not able to generate high-quality sign language videos, as each mainstream human image synthesis approach has its own limitations. Currently, the mainstream human image synthesis methods can be broadly categorized into two classes: Pose Transfer\cite{ma2017pose} and Image Animation\cite{siarohin2019first}: 

Pose Transfer methods\cite{Stoll_ACVR2020,saunders2022signing,fang2023signdiff} typically use Generative Adversarial Networks (GANs)\cite{goodfellow2014generative} to fuse human source images with specified driving poses to generate driving pose images. However, it is difficult for them to ensure appearance coherence when generating video frames guided by different poses, which leads to the problem of poor temporal consistency. For example, in the left half of \cref{fig:problem}, the clothing between frames changes from short-sleeved to long sleeves.

Image Animation methods\cite{suo2022jointly,siarohin2021motion,zhao2022thin,siarohin2019first} learn the motion representation from optical flow. They unsupervisedly learn the keypoints of the motion on the source and driving images, generate the optical flow by the changes of the keypoints, and the features of the source image are warped by the optical flow to generate the target image. It is difficult for them to generate fine details in sign language videos, because they focus on the macro pose of the human body. 
The details of complex regions such as the hands in the generated video tend to be blurred, distorted, and structurally incorrect, For example, in the right half of \cref{fig:problem}, the facial expressions and hand structures of the pictures generated by image animation method are distorted.

\begin{figure}[tb]
  \centering
  \includegraphics[width=0.9\textwidth]{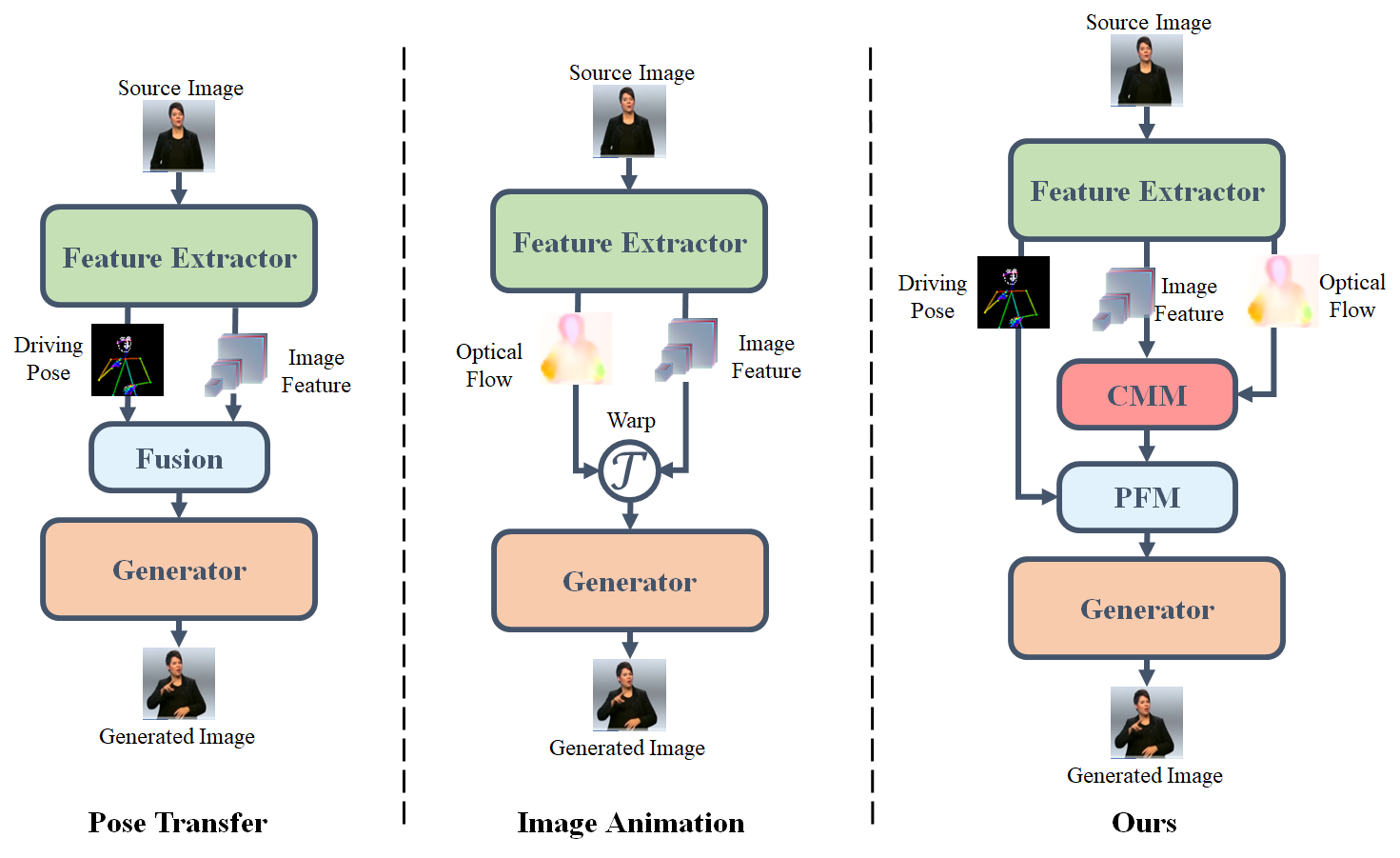}
  \caption{
    Comparison of previous two types of methods and the proposed method.
  }
  \label{fig:compare}
\end{figure}

To overcome these limitations, we propose a novel framework, called Pose-Guided Motion Model (PGMM), for generating fine-grained sign language videos using human pose information. A comparison with the above two classes of approaches is shown in \cref{fig:compare}. First, we introduce a new Coarse Motion Module (CMM), which completes the deformation of features by optical flow warping, thus completing the motion of coarse-grained structures without changing the details. Then, we propose a new Pose Fusion Module (PFM), which guides the local fusion of features by the pose information to refine the local fusion of features, thus completing the fine-grained generation. In addition, we design additional loss functions to improve the detail generation. Finally, to highlight the temporal consistency requirement in SLVG, we design a new Temporal Consistency Difference (TCD) metric, which quantitatively evaluates the degree of temporal consistency of a video by comparing the difference between the frames from the reconstructed video and the front and back frames from the target video.

The main contributions of this paper can be summarized as follows:
    \begin{itemize}
        \item We propose a novel sign language video generation framework that generates sign language videos with temporal consistency and realistic details by decoupling the sign language video generation task into two parts: coarse-grained structural motion via optical flow warping, and fine-grained generation via pose guidance. 
        \item We present a new Pose Fusion Module (PFM), which effectively fuses the pose modality information and image modality information using one Cross Attention mechanism to generate fine-grained detail regions such as the human hand and face.
        \item Qualitative and quantitative experiments extensively demonstrate that our method generates videos with good temporal consistency and details, and achieves state-of-the-art performance on multiple sign language datasets. Visualization results also show the effectiveness of our proposed modular partitioning.
    \end{itemize}

\section{Related Work}
\subsection{Sign Language Production(SLP)}
Sign Language Production (SLP), the process of translating spoken language into sign language currently relies on various methods to express sign language, including 2D pose sequences \cite{saunders2021continuous,saunders2021skeletal,saunders2021mixed}, 3D avatar sequences \cite{baltatzis2023neural,cox2002tessa,zuo2024simple,mcdonald2016automated,kipp2011sign}, or real-life sign language videos \cite{kissel2021pose,cui2019deep,saunders2020everybody,stoll2020text2sign,saunders2022signing,fang2023signdiff}. Real-life sign language videos are considered the most comprehensible form of sign language expression for signers due to their fidelity in realism \cite{ventura2021everybody}.

\subsection{Human Image Synthesis}

\subsubsection{Pose Transfer.}
This method was utilized in SLVG works such as \cite{Stoll_ACVR2020,saunders2022signing,fang2023signdiff}, was first proposed by \cite{ma2017pose}. Its synthesis process is guided by the human \sloppy pose condition, which usually represented as human pose keypoints coordinates. This method generates realistic human images with fine-grained hand details, due to the precise target pose. Current methods primarily rely on flow \cite{ren2020deep,liu2019liquid,li2019dense,tang2021structure,ma2021fda,ren2021combining} or attention \cite{zhang2022exploring,ren2022neural,tang2020xinggan,li2020pona,liu2022dynast}. 

Although these methods provide rich detail through precise pose guidance, they generate videos with poor temporal consistency due to their focus on image-based generation and inadequate decoupling of the appearance and the motion of the human body in the latent space.

\subsubsection{Image Animation.}
This method was utilized in \cite{suo2022jointly}. It was first proposed by \cite{siarohin2019animating} and developed by a series of work \cite{siarohin2019first,siarohin2021motion,wang2022latent,tao2022structure,zhao2022thin}. the method learns to detect keypoints in an unsupervised way to generate optical flow, and it utilizes optical flow information to generate dynamic and animated sequences of images. 

These methods has better decoupling of human appearance and motion, thus temporal consistency is better. However, the disadvantage of this method is that the transformations represented by the optical flow are too coarse to generate local details and fine motion.

\subsection{Pose\&RGB Feature Fusion}
Previously, in Action Recognition, Sign Language Recognition, and Sign Language Translation tasks, the recognition/translation ability is often improved by fusing Pose and RGB features. The simplest way \cite{bruce2022mmnet,jiang2021skeleton} can be summing the prediction results after the classifier of RGB branch and pose branch. \cite{chen2022two} directly sum the RGB and pose feature in convolution networks. \cite{ahn2023star,kim2023cross} concatenate the input RGB and pose token to the transformer, \cite{das2020vpn,das2021vpn++} use GCN aggregation with the pose graph to compute the spatio-temporal attention weights. 

In this paper, we proposed a novel Pose\&RGB feature fusion method suitable for generating fine-grained sign language videos.

\section{Methodology}

\begin{figure}[tb]
  \centering
  \includegraphics[width=\textwidth]{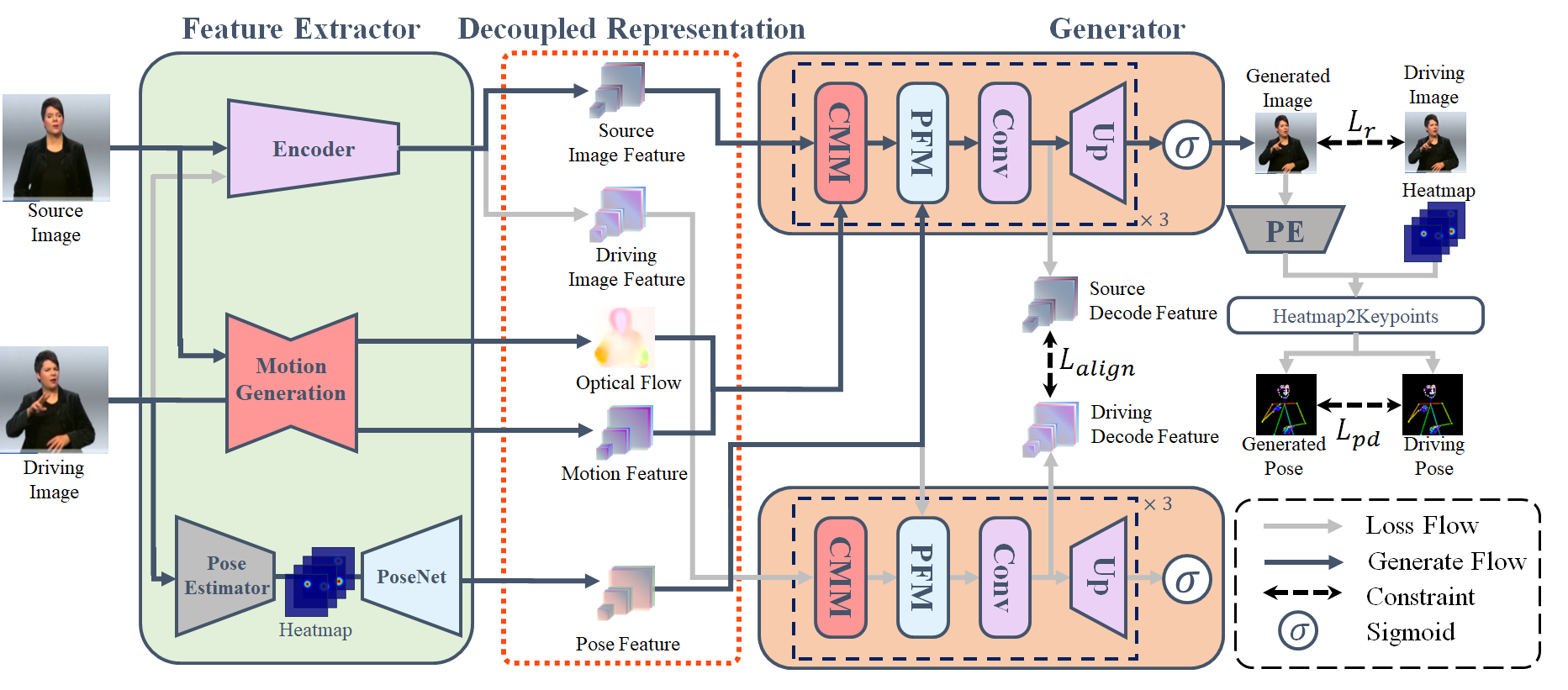}
  \caption{The framework of the proposed method. The decoupled representations are first extracted by Feature Extractor. It consists of RGB features, motion features, and pose features. The Generator then uses the above features to perform the image generation, which consists of multiple sets of a Coarse Motion Module for structural motion, a Pose Fusion Module for generating details through pose-guided fusion, a Convolution Module for image inpainting, and an Up Convolution Module for up-sampling features.
  }
  \label{fig:frame}
\end{figure}

The core idea of our approach is to decouple the representation of a sign language image $I$ into the human appearance $X_A \in \mathbb{A}$, the coarse-grained motion $X_M \in \mathbb{M}$ and the fine-grained detail $X_D \in \mathbb{T}$, i.e. , to obtain the mapping $g$ such that $g(X_A,X_M,X_D)= I$. 
\cref{fig:frame} shows the overview of our proposed framework. It generates the reconstructed driving image $\hat{I}_d$ given a source image $I_s$ and a driving image $I_d$. It can be expressed as:
\begin{equation}
    \hat{I}_d = {\rm PGMM}(I_s, I_d).
    \label{eq:1}
\end{equation}
The framework consists of the following modules:
\begin{itemize}
    \item \textbf{Encoder} extracts appearance representation via input image.
    \item \textbf{Motion Generation} extracts motion representation via the source image and the driving image to guide the motion transfer.
    \item \textbf{Pose Estimator} is a pre-trained human Pose Estimator, to extract driving image's pose heatmap.
    \item \textbf{PoseNet} extracts detail representation via pose heatmap to guide the details generation.
    \item \textbf{Generator} generates the image by decoupled representations. It consists of multiple sets of a Coarse Motion Module, a Pose Fusion Module, a Convolution Module, and an Up-sampling Module.
    \item \textbf{Coarse Motion Module} (CMM)  completes the coarse motion transfer.
    \item \textbf{Pose Fusion Module} (PFM) generates fine-grained details.
\end{itemize}
    \subsection{Feature Extractor}
    \label{sec:FE}
    In our Feature Extractor, we designed three branches: image, motion, and pose, to extract the decoupled representations $X_s,X_m,X_p$ from input image $I \in \mathbb{R}^{H \times W \times 3}$ respectively. 
    
    In image branch, we use Encoder $E_I$ to extract the source image feature $X_s \in \mathbb{A}$, as $X_s = E_I(I_s)$.
    After an initial $7 \times 7$ convolutional layer with a stride of $3$ that increases the feature channels, the encoder uses three down-sampling blocks to gradually extract high-level features from the input image.
    Every down-sampling block consists of a $3 \times 3$ convolutional layer with a stride of $1$, an instance normalization layer, a ReLU activation layer, and a $2 \times 2$ average pooling layer with a stride of $2$.
    
    In motion branch,  we use Motion Generation $E_M$ to extract the $I_s \rightarrow I_d$ motion feature $X_m \in \mathbb{M}$ as $X_m = E_{M}(I_s,I_d)$.
    Like \cite{zhao2022thin}, $E_M$ consists of a keypoints detector and a dense motion, where keypoints detector receives $I_s, I_d$, outputs the respective motion keypoints $K_s, K_d$, dense motion receives $K_s, K_d \in \mathbb{R}^{K \times 2}$ and $I_s$, constructs transformations on $K_s \rightarrow K_d$ and extracts the motion features step-by-step through an Hourglass network. 
    
    In pose branch, we use Pose Estimator $P_e$ and PoseNet $P_d$ to extract the pose feature $X_p \in \mathbb{T}$, as $X_p = P_d(P_e(I_d))$. $P_e$ is a human pose heatmap estimator pre-trained with freezing parameters, it receives image $I$ to estimate heatmap $H \in \mathbb{R}^{P \times H_h \times W_h}$, where $P$ is the number of poses, $H_h,W_h$ are the height and width of the heatmap. 
    PoseNet consists of an interpolation and three up-sampling blocks. It first resizes the heatmap to the size of the biggest scale image feature and then gradually extracts fine-grained pose features. Every up-sampling block consists of a $3 \times 3$ transposed convolutional layer with a stride of $2$, an instance normalization layer, and a ReLU activation layer.


    \subsection{Generator}
    We construct the generator $G$ to approximate  mapping $g$, we use the source image $I_s$ and the driving image $I_d$ of the same person to reconstruct driving image $\hat{I}_d$ by combining the source image's appearance $X_s$, the motion $X_m$ between $I_s$ and $I_d$ the driving image's fine-grained pose $X_p$, we simplify \cref{eq:1}:
    \begin{align}
        \hat{I}_d &= G(X_s, X_m, X_p), \notag \\
        \hat{I}_d &= G(E_I(I_s), E_m(I_s,I_d), P_d(P_e(I))). \label{eq:2}
    \end{align}
    Here, $G$ represents the Generator. It consists of multiple sets of Coarse Motion Module, Pose Fusion Module, Convolution Module, and Up-sampling module, and a sigmoid at the end.
    Every Convolution Module consists of 2 ResBlock for inpainting. Every Up-sampling module consists of an interpolation($2\times$), a $3 \times 3$ convolutional layer with a stride of $1$, an instance normalization layer, and a ReLU activation layer.
    \label{sec:G}
\begin{figure}[tb]
  \centering
  \includegraphics[width=0.75\textwidth]{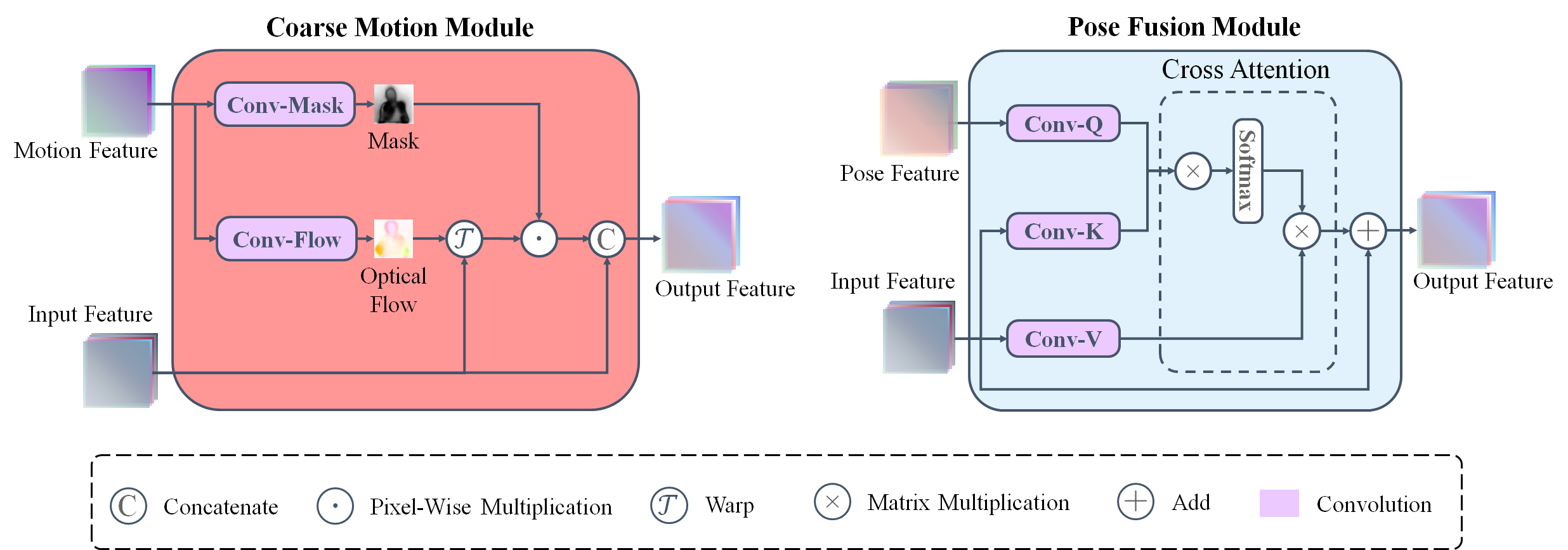}
  \caption{
    The structure of our Coarse Motion Module (CMM) and our Pose Fusion Module (PFM).
  }
  \label{fig:modules}
\end{figure}
        \subsubsection{Coarse Motion Module.}
        Coarse Motion Module (CMM) completes the deformation of features by optical flow warping, thus completing the motion of coarse-grained structures without changing the appearance. 
        
        As shown in \cref{fig:modules}, firstly, the motion features $X_m$ are fed into Conv-Flow module $F$ to calculate the motion optical flow $F_m \in \mathbb{R}^{H_f \times W_f \times 2}$, as: $F_m = F(X_m)$. Conv-Flow module consists of a 7 × 7 convolutional layer with a stride of $1$ and a softmax.
        Secondly, the input image features $X_{in} \in \mathbb{R}^{H_{in} \times W_{in} \times C_{in}}$ are warped $\mathcal{T}$ by the motion optical flow $F$ to obtain the image features $X'_{in} \in \mathbb{R}^{H_{in} \times W_{in} \times C_{in}}$ corresponding to the input image features after the coarse motion transformation, and then, input the motion features $X_m\in \mathbb{R}^{H_{in} \times W_{in} \times C_{in}}$ to Conv-Mask module $M$, which consists of a $3 \times 3$ convolutional layer with a stride of $1$ and a softmax, thus obtaining the mask of the occluded region in the motion $O \in \mathbb{R}^{H_{in} \times W_{ in}}$, correspondingly, the mask of the appearance region that needs to be supplemented after the motion transformation is $1-O$. Then concatenate the original appearance feature $X_{in}$ and pixel-wise multiplication of $1-O$ and $X'_{in}$to obtain the output feature $X_{out} \in \mathbb{R}^{H_{in} \times W_{in} \times (2 \times C_{in})}$, as:
        \begin{equation}
            X_{out} = [X_{in},(1-M(X_m))\odot\mathcal{T}(F(X_m),X_{in})],
        \end{equation}
        where notion $\odot$ is pixel-wise multiplication.
        \subsubsection{Pose Fusion Module.}
        Pose Fusion Module (PFM) guides the local fusion of features by pose information, thus completing the fine-grained generation. Then, we design additional loss functions to improve detail generation.
        As shown in \cref{fig:modules}, PFM receives input image feature $X_{in} \in \mathbb{R}^{H_{in} \times W_{in} \times C_{in}}$ and fine-grained pose feature $X_{p} \in \mathbb{R}^{H_{in} \times W_{in} \times C_{in}}$, completes the fusion of $X_{in}$ and $X_p$ through the cross-attention mechanism. Firstly, three convolution modules $Q,K,V$ with the same structure but different parameters are used to obtain the Query, Key, and Value needed for attention. Every convolution module consists of a $3 \times 3$ convolutional layer with a stride of $1$, an instance normalization layer, and a ReLU activation layer. Secondly, the detailed feature $X'_{in} \in \mathbb{R}^{H_{in} \times W_{in} \times C_{in}}$ is computed by cross-attention ${\rm Atten}$ from Query, Key, and Value. Finally, we add $X'_{in}$ to $X_{in}$ by summation to generate image feature $X_{out} \in \mathbb{R}^{H_{in} \times W_{in} \times C_{in}}$, as:
\begin{align}
            X_{out} &= X_{in} + X'_{in},  \\
            X_{out} &= X_{in} + {\rm Atten}(Q(X_p),K(X_{in}),V(X_{in})),  \\
            X_{out} &= X_{in} + {\rm softmax}(\frac{Q(X_p)K(X_{in})^T}{\sqrt{C_{in}}})V(X_{in}).
\end{align}

    \subsection{Training and Inference}
    
    In the training stage, we randomly select any two frames from the same video as the source image $I_s$ and driving image $I_d$. We need to extract the image feature $X_s,X_d$ of $I_s$ and $I_d$ through Encoder, the motion feature $X_m$ of $I_s,I_d$ extracted through Motion Generation, and the pose feature $X_p$ of $I_d$ to reconstruct $I_d$ by generated image $\hat{I}_d$.
    The loss function is a combination of multiple losses:
    \begin{equation}
        L = L_r + L_{pd} + L_{align},
    \end{equation}
    where $L_{pd}$ is the Pose Distance Loss, $L_{align}$ is the Feature Alignment Loss, and $L_r$ is the loss function in TPSMM\cite{zhao2022thin}, consisting of perceptual loss, equivariance loss, background loss, and warp loss.
        \subsubsection{Pose Distance Loss.}
        We propose the Pose Distance Loss to enable the model to fully exploit the semantic information of the pose in the learning process and improve the detail generation capability of the model. It is calculated as shown in \cref{fig:frame}, as:
        \begin{equation}
            L_{pd} = \frac{1}{K} \sum_{i=1}^{K} | {k_i(P_e(\hat{I}_d))} - {k_i(H)} |,
        \end{equation}
        where $k_i$ gets the coordinate of the point that corresponds to the maximum value of the $i^{th}$ pose of the heatmap. 
        \subsubsection{Feature Alignment Loss.}
            To better decouple the $\mathbb{A},\mathbb{M},\mathbb{T}$ space, we align the features $Y_s$ and $Y_d$ in $G$ by:
        \begin{align}
            L_{align} &= |Y_s - Y_d|, 
        \end{align}
        where $Y_s$ and $Y_d$ correspond to source decode feature and driving decode feature in \cref{fig:frame}, which are obtained via:
\begin{align}
Y_s^i &= G^i(X_s, X_m, X_p), \\
Y_d^i &= G^i(X_d, \emptyset, X_p).
\end{align}
        Here $i$ denotes the output of the $i^{th}$ convolution module in Generator, and $\emptyset$ denotes the default, i.e., features are not transformed by CMM in Generator. The actual meaning of $Y_s, Y_d$ is the intermediate variable when using the features of $I_s$ and $I_d$ to complete the reconstruction via Generator, since the reconstructions have the same goal, the distances in the space of their variables should also converge to 0.
        \subsubsection{Inference.}
        For inference, the generator is fed with the image features of the source image, motion features extracted through the motion generator, and pose features of each frame of the driving video, allowing the character in the source image to perform the actions in the driving video. Influenced by CMM and PFM, we can obtain fine-grained videos with good temporal consistency.

\section{Experiments}
\subsection{Experimental Setup}
        \subsubsection{Datasets.} The benchmark datasets we tested in our experiments are as follows:
        \begin{itemize}
            \item LSA64\cite{ronchetti2023lsa64}: It is a small-scale dataset of Argentinian Sign Language (LSA). It comprises 3,200 sign language videos performed by 10 different signers, encompassing a total of 64 unique words. For our experiments, we select word categories numbered 57-64 as the test set, other 56 categories serve as the training set. Consequently, there are a total of 2,800 training videos and 400 test videos.
            \item WLASL-2000\cite{li2020word}: It is a large-scale American Sign Language (ASL) dataset consisting of 21,083 videos performed by over 100 signers, covering a vocabulary of 2,000 words. We utilize the official train-test split, with 14,289 videos for training and 2,878 videos for testing.
            \item RWTH-PHOENIX-Weather 2014T\cite{Phoenix14t}: It is a German sign language dataset of German weather forecast broadcasts. It contains 8,247 sentences with a vocabulary of 1,085 signs, split into 7,096 training instances, 519 development (Dev) instances and 642 testing (Test) instances.
            \item CSL-Daily\cite{zhou2021improving}: This large-scale Chinese sign language dataset revolves the daily life, recorded by 10 signers. It contains 20,654 sentences, divided into 18,401 training samples, 1,077 development (Dev) samples and 1,176 testing (Test) samples.
        \end{itemize}
        \subsubsection{Metrics.}
        For evaluating the image quality of model reconstructions, we adopted L1\cite{L1}, Structural SIMilarity (SSIM)\cite{SSIM}, Learned Perceptual Image Patch Similarity (LPIPS)\cite{LPIPS} following previous work. For the video-based evaluation, we utilized Fréchet Video Distance (FVD)\cite{FVD}, Word Error Rate (WER)\cite{WER}, and our proposed Temporal Consistency Difference (TCD) as evaluation metrics to compare the reconstructed video with the ground truth.
        \begin{itemize}
            \item Fréchet Video Distance (FVD)\cite{FVD}: It is a metric similar to FID\cite{FID} but specifically designed for assessing the quality of generated videos, where we employ a pre-trained I3D\cite{I3D} model for feature extraction.
            \item {Temporal Consistency Difference (TCD): To measure the degree of temporal consistency of the video, we propose TCD.
            It is calculated by the difference $D$ between the generated frame $\hat{I}_t$ and the average values of the real video frames before and after the corresponding frame:
            \begin{equation}
                D =  |\hat{I}_t- \frac{1}{2} (I_{t-1} + I_{t+1})|.
            \end{equation}
            And calculate the per-frame average of the proportion of pixels whose difference exceeds the set threshold $T$. 
            \begin{equation}
                TCD = \frac{1}{N-2} \sum_{t=1}^{N-2} \frac{1}{H \times W} \sum_{h=1}^{H} \sum_{w=1}^{W} [ D(h,w) > T ].
            \end{equation}
            }
            \item Word Error Rate (WER)\cite{WER}: It is defined as the minimal summation of the substitution, insertion, and deletion operations to convert the predicted sentence to the reference sentence:
            \begin{equation}
                WER = \frac{\#sub + \#ins + \#del}{\#reference}.
            \end{equation}

        \end{itemize}
    \subsubsection{Inplementation Details.}
    For Motion Generation, we use the TPSMM\cite{zhao2022thin} approach using ResNet-18 as the visual backbone to generate 100 keypoints in groups of 5 keypoints to generate 20 groups of TPS transforms.
    For Pose Estimator, we used HRNet-W32 pre-trained on COCO-WholeBody\cite{HRNet}.We selected 36 heatmaps of pose points.
    For evaluation, in TCD, we set $T=0.5$. In WER, we used CorrNet\cite{hu2023continuous} for sign language recognition, and we use checkpoints provided by the authors, trained on the Phoenix-2014T and CSL-Daily datasets at 256*256 resolution.

    For all datasets, we crop and resize input videos to 128*128. For each video, we randomly select five pairs of source and driving images. For all experiments, we use the Adam optimizer with an initial learning rate of $2 \times 10^{-4}$, which is decreased by a factor of 10 at the 60th and 90th epochs. We train the entire network for 100 epochs on one NVIDIA GeForce 3090 GPU.

\subsection{Comparison with state-of-the-art Methods}
We compared our method with state-of-the-art methods including DynaST\cite{liu2022dynast}, MRAA\cite{siarohin2021motion}, TPSMM\cite{zhao2022thin}.
DynaST is a pose-guided human image synthesis \sloppy method, that generates images guided by human body pose-points. MRAA is a region-based method that estimates human joint regions in an unsupervised manner. TPSMM is a keypoints-based method that estimates human body keypoints in an unsupervised manner, similar to our approach. 
We evaluate them on video reconstruction and cross-person video generation tasks.
Video reconstruction is the task of reconstructing the original video by inputting one frame of the video as the source image and the rest of the frames as the driving images. 
Cross-person video generation is the task of taking different two persons as the source image and the driving image as input respectively. It can effectively test the general applicability of SLVG methods. (Please refer to the supplementary materials for more experimental results.)

\begin{figure}[tb]
  \centering
  \includegraphics[width=1\textwidth]{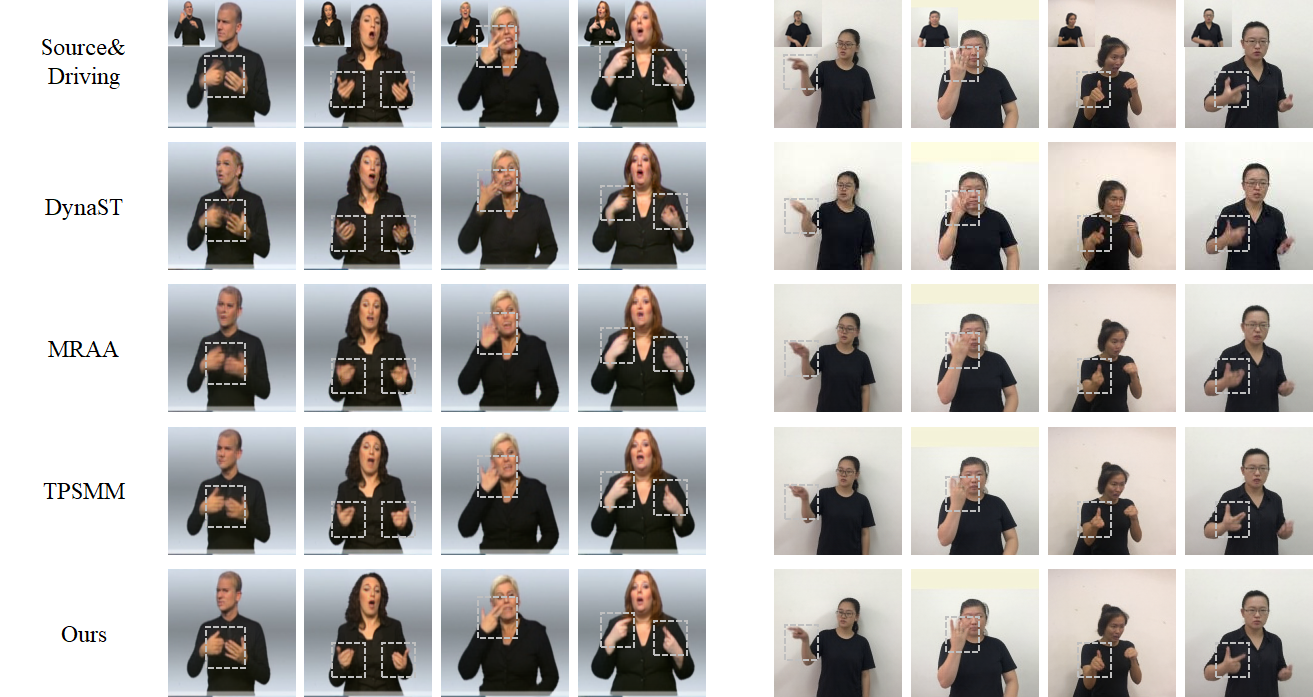}
  \caption{
    Qualitative comparison with DyanST\cite{liu2022dynast}, MRAA\cite{siarohin2021motion}, and TPSMM\cite{zhao2022thin} on video reconstruction: Phoenix-2014T(left) and CSL-Daily(right).
  }
  \label{fig:recons}
\end{figure}

\subsubsection{Video Reconstruction.}
For DynaST, to explore its best performance, we use all 133 key points annotated in COCO-WholeBody as input guidance. For MRAA and TPSMM, to be fair, we select the same number of transformations $K=20$ (number of key regions, number of TPS). We evaluate the video reconstruction capabilities of all methods using the image quality and video quality evaluation metrics mentioned earlier. As shown in \cref{tab:ISLR}, \ref{tab:Phoenix} and \ref{tab:CSL}, our proposed method surpasses the others, achieving state-of-the-art performance in terms of all metrics on three datasets.
\begin{table}[tb]
\caption{The quantitive evaluation results for video reconstruction on LSA64 dataset and WLASL-2000 dataset.
}
\label{tab:ISLR}
\fontsize{7pt}{8pt}
\selectfont
\centering
\begin{tabular}{c|ccc|cc|ccc|cc}

\toprule
\multicolumn{1}{c}{\multirow{2}{*}{Method}} & \multicolumn{5}{c}{LSA64}   & \multicolumn{5}{c}{WLASL-2000} \\ 
\multicolumn{1}{c}{}                        & L1↓ & SSIM↑ & LPIPS↓ & FVD↓ & TCD↓ & L1↓  & SSIM↑  & LPIPS↓  & FVD↓ & TCD↓ \\ \midrule
MRAA                                        &0.01196&0.9324&0.01996&     184.668&       0.150
&0.01285&0.9384&0.02300&      289.722&        0.066
\\
TPSMM                                       &0.01342&0.9208&0.02261&      182.785&      0.130
&        0.0110&         0.9501&   \bf 0.01685& 225.109&   \bf 0.064\\ \midrule
\bf Ours                                        &\bf 0.01144&\bf 0.9395&\bf 0.01638&\bf 154.726&\bf 0.125&\bf 0.01085&\bf 0.9513&0.01695& \bf 199.095& 0.067\\ \bottomrule
\end{tabular}
\end{table}

\begin{table}[tb]
\caption{The quantitative evaluation results for video reconstruction on PHOENIX-2014T dataset.
}
\label{tab:Phoenix}
\fontsize{6pt}{8pt}
\renewcommand{\arraystretch}{0.8}
\setlength{\tabcolsep}{0.9pt}
\selectfont
\centering
\begin{tabular}{c|ccc|ccc|ccc|ccc}
\toprule
\multicolumn{1}{c}{\multirow{2}{*}{Method}} & \multicolumn{6}{c}{PHOENIX-2014T(dev)}   & \multicolumn{6}{c}{PHOENIX-2014T(test)} \\ 
\multicolumn{1}{c}{}                        & L1↓ & SSIM↑ & LPIPS↓ & FVD↓ & TCD↓ & WER↓ & L1↓  & SSIM↑  & LPIPS↓ & FVD↓ & TCD↓ & WER↓ \\ \midrule
DynaST                                        &0.03965&0.8113&0.05748
&     665.054
&       0.424
&68.5\%&0.03938&0.8093&      0.05758
&        626.207
& 0.436
&64.8\%\\
MRAA                                        &0.01752&0.9110&0.03001
&     300.649
&       0.118
&69.5\%&0.01718&0.9117&      0.02980&        303.350& 
0.116
& 70.7\%\\
TPSMM                                       &0.01519&0.9276&0.02244
&      202.535
&      0.120
&        56.0\%&         0.01487&      0.9279& 0.02254
& 209.297
& 
0.116
& 55.2\%\\ \midrule
\bf Ours                                        &\bf 0.01405&\bf 0.9376&\bf 0.01853&\bf 124.251&\bf 0.117&\bf 39.7\%&\bf 0.01380&\bf 0.9377& \bf 0.01883& \bf 136.472& \bf 0.113&\bf 39.7\%\\ \bottomrule
\end{tabular}
\end{table}

Additionally, we visualize a set of reconstruction results, as shown in \cref{fig:recons}. DynaST\cite{liu2022dynast} exhibits structural distortions in fine finger details, and there is a semantic inconsistency between frames (e.g., collar shape, mouth shape). Both MRAA\cite{siarohin2021motion} and TPSMM\cite{zhao2022thin} demonstrate better frame consistency than DynaST. However, MRAA suffers from severe blurriness and distortion in finger parts, which we attribute to the large size of predicted key regions, and insufficient to capture detailed motion information. Meanwhile, TPSMM also loses details in hand parts, indicating the current unsupervised methods' shortcomings in detail generation capabilities. Our method excels in semantic consistency between frames and the quality of facial expressions, finger details, etc., compared to other methods. This result validates our method's ability to generate fine-grained sign language videos by incorporating human body structure pose-points. We achieve this by blending the added pose-points with RGB features during the decoding process using PFM, guiding the generation of image details. For example, adding facial and hand pose-points information assists our method in generating specified expressions, mouth shapes, and finger details.

\subsubsection{Cross-Person Video Generation.}

\begin{figure}[tb]
  \centering
  \includegraphics[width=\textwidth]{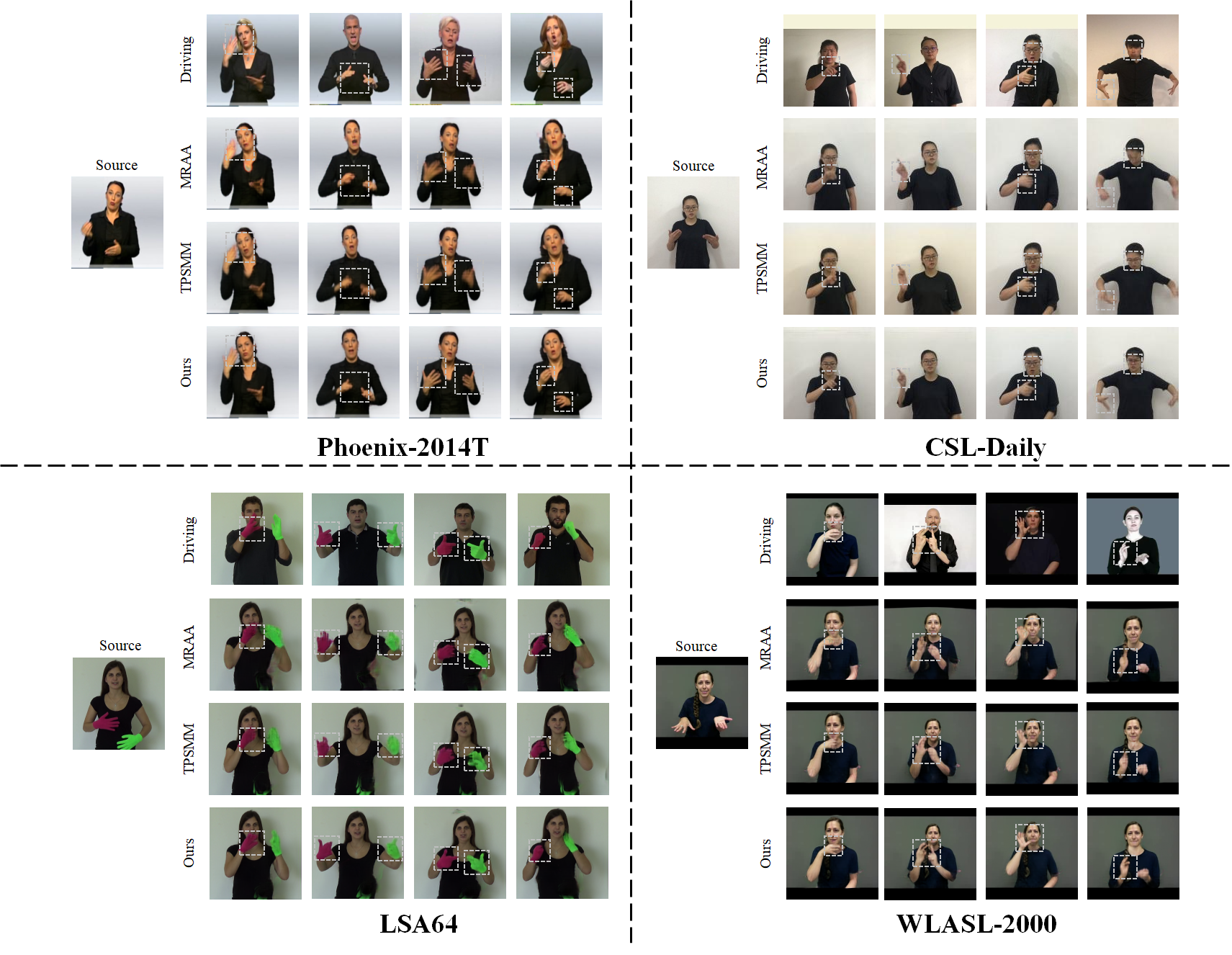}
  \caption{
    Qualitative comparison with MRAA\cite{siarohin2021motion} and TPSMM\cite{zhao2022thin} on cross-person video generation task.
  }
  \label{fig:cross}
\end{figure}

\cref{fig:cross} shows selected Cross-Person Video Generation results of our method compared with MRAA and TPSMM. For expression details, most of the images generated by MRAA and TPSMM are not accurate enough, such as head angle and lip shape. For hand details, the images generated by MRAA usually lack finger details, and the images generated by TPSMM can only generate specific finger details in rare cases. , while our method can generate accurate expressions and hand details. However, when generating samples with large differences between the source image's hairstyle and the driving image's hairstyle, although our method can generate accurate expressions, the hairstyle semantics cannot be well maintained (see \cref{fig:cross} Phoenix-2014T and CSL- Daily’s 4th example).

\subsection{Ablation Study}

\begin{table}[tb]
\caption{The quantitative evaluation results for video reconstruction on CSL-Daily dataset.
}
\label{tab:CSL}
\fontsize{6pt}{8pt}
\renewcommand{\arraystretch}{0.8}
\setlength{\tabcolsep}{0.9pt}
\selectfont
\centering
\begin{tabular}{c|ccc|ccc|ccc|ccc}
\toprule
\multicolumn{1}{c}{\multirow{2}{*}{Method}} & \multicolumn{6}{c}{CSL-Daily(dev)}   & \multicolumn{6}{c}{CSL-Daily(test)} \\ 
\multicolumn{1}{c}{}                        & L1↓ & SSIM↑ & LPIPS↓ & FVD↓ & TCD↓ & WER↓ & L1↓  & SSIM↑  & LPIPS↓ & FVD↓ & TCD↓ & WER↓ \\ \midrule
DynaST                                        &0.04202&0.8506&0.04161
&479.937
&0.655
&94.0\%&0.04253&0.8485&0.04249
&478.063
&0.654
& 93.9\%\\
MRAA                                        &0.01363&0.9417&0.01643
&     137.293
&       0.223
&71.5\%&0.01373&0.9410&      0.01668
&        136.534
& 0.225
& 71.1\%\\
TPSMM                                       &0.01113&0.9525&0.01194
&      99.119
&      0.192
&        60.4\%&         0.01117&      0.9520& 0.01205
& 98.778
& 0.192
&59.7\%\\ \midrule
\bf Ours                                        &\bf 0.01092&\bf 0.9536&\bf 0.01113&\bf 83.670&\bf 0.176&\bf 57.5\%&\bf 0.01098&\bf 0.9532& \bf 0.01123& \bf 82.974& \bf 0.176&\bf 57.1\%\\ \bottomrule
\end{tabular}
\end{table}

\begin{table}[tb]
\caption{The ablation study on the proposed Pose\&RGB fusion method and loss constraint on PHOENIX-2014T dataset.
}
\label{tab:Ablation}
\fontsize{5pt}{6pt}
\renewcommand{\arraystretch}{0.8}
\setlength{\tabcolsep}{0.9pt}
\selectfont
\centering
\begin{tabular}{c|ccc|ccc|ccc|ccc}
\toprule
\multicolumn{1}{c}{\multirow{2}{*}{Method}} & \multicolumn{6}{c}{PHOENIX-2014T(dev)}   & \multicolumn{6}{c}{PHOENIX-2014T(test)} \\ 
\multicolumn{1}{c}{}& L1↓ & SSIM↑ & LPIPS↓ & FVD↓ & TCD↓ & WER↓ & L1↓  & SSIM↑  & LPIPS↓ & FVD↓ & TCD↓ & WER↓ \\ \midrule
baseline&0.01519&0.9276&0.02244
&202.535
&0.120&56.0\%&0.01487&0.9279&0.02254
&209.297
&0.116&55.2\%\\
+pose&0.01443&0.9325&0.2007&      153.94&      0.118&        44.1\%&         0.01426&      0.9325& 0.2033& 140.957
&  0.117& 44.0\%\\ 
+PFM&0.01420&0.9365&0.01891
&      124.989
&      0.122&        40.4\%&         0.01396&      0.9366& 0.01916
& 135.244
&  0.126& 40.1\%\\
+PFM,$L_{p}$&0.01419&0.9372&0.0187
&      124.470&      \bf 0.102&        40.0\%&         0.01396&   0.9374& 0.01895
& \bf 133.178&  \bf 0.105&  \bf 39.4\%\\
+PFM,$L_{align}$&\bf 0.01400&0.9374&0.01858
&      128.404
&      0.115&        41.2\%&         \bf 0.01375&   0.9376&  0.01887
& 140.957
&  0.119& 40.6\%\\
\midrule
\bf Full Model &0.01405
&\bf 0.9376&\bf 0.01853&\bf 124.251&0.117&\bf 39.7\%&0.01380&\bf 0.9377&\bf 0.01883&136.472&0.113&39.7\%\\
\bottomrule
\end{tabular}
\end{table}
We choose the TPSMM\cite{zhao2022thin} model as the baseline and compare it with several other models: +pose, which directly adds the pose features to the RGB features like \cite{chen2022two} of mixing multimodal information; +PFM, which utilizes our proposed multimodal mixing method PFM; +PFM,$L_p$, which adds Pose Distance Loss as a constraint in addition to PFM; and +PFM,$L_{align}$, which adds Feature Alignment Loss as a constraint in addition to PFM.

The results are shown in \cref{tab:Ablation}. From experiment results, we observe that due to the employment of the Cross Attention mechanism, using PFM as the multimodal mixing method can better extract the regions from the pose that play a crucial role in the details of RGB, thus achieving better fusion of Pose and RGB features compared to the +pose method. Regarding the two constraints proposed in the paper, we find that Pose Distance Loss, due to its constraint on abstract information, performs better on video evaluations, whereas Feature Alignment Loss, as it fully utilizes the detailed information of the source image during training, performs better on L1.

\subsubsection{Effect of CMM and PFM.}
\begin{figure}[tb]
  \centering
  \includegraphics[width=1\textwidth]{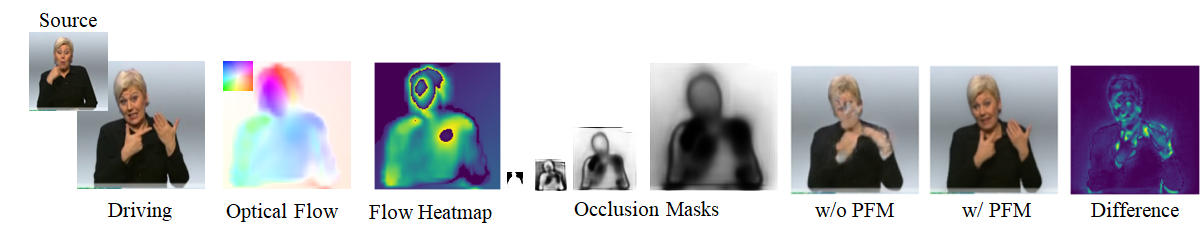}
  \caption{
    Visualizations of intermediate variables of CMM and PFM.
  }
  \label{fig:vis}
\end{figure}
We visualized some intermediate variables in our method to demonstrate the division of labor between CMM and PFM, as shown in \cref{fig:vis}. For the structural motion of coarse-grained structures, the optical flow generated by Motion Generation is mainly large-scale actions, which we visualize as heatmaps (blue corresponds to lower values and yellow is higher in the figure). The areas with higher optical flow include the arms and the head. For the occlusion masks generated by CMM (white represents the occluded area, while optical flow changes the unobstructed black area), we can see that the areas changed by optical flow in the CMM module include the head and arms, and CMM's precision is coarse. For fine-grained generation, we removed the PFM module for generation and calculated the difference with the results generated by the model without PFM removal. The difference was visualized as a heatmap. It can be seen that the PFM module is mainly responsible for generating contours and details in areas such as facial features and fingers. 

\section{Potential Negative Social Impact}
Our method can be utilized to create convincingly fake videos or images, which may lead to the mass dissemination of false information, thus affecting the public's understanding and judgment of authenticity. In addition, the sign language videos we generate may be misinterpreted. Therefore, in critical application scenarios, users should verify the accuracy of sign language.

\section{Conclusion and Future Work}
In this paper, we proposed a framework for sign language video generation, called Pose Guided Motion Model (PGMM). Unlike existing studies, PGMM uses human pose to generate sign language videos with fine details. We also present one Coarse Motion Module (CMM) and one Pose Fusion Module (PFM) for coarse-grained motion generation and fine-grained detail generation, respectively, to better fuse the human pose information into the details of the image while maintaining temporal coherence. We propose Pose Distance Loss and Feature Alignment Loss, which enable the model to fully exploit the semantic information of the pose in the learning process and improve the detail generation capability of the model. In addition, considering the coherence requirement of sign language video generation, we designed the quantitative metric, Temporal Consistency Difference (TCD), to measure the temporal consistency of the video. A large number of qualitative and quantitative experiments demonstrated that our method can generate sign language videos with better details and temporal consistency compared with existing works. 

In the future, we will continue to explore how the video generation capabilities of this method can be leveraged for Sign Language Production (SLP) tasks.

\bibliographystyle{splncs04}

\begin{thebibliography}{10}
\providecommand{\url}[1]{\texttt{#1}}
\providecommand{\urlprefix}{URL }
\providecommand{\doi}[1]{https://doi.org/#1}

\bibitem{ahn2023star}
Ahn, D., Kim, S., Hong, H., Ko, B.C.: Star-transformer: a spatio-temporal cross attention transformer for human action recognition. In: The IEEE Winter Conference on Applications of Computer Vision. pp. 3330--3339 (2023)

\bibitem{baltatzis2023neural}
Baltatzis, V., Potamias, R.A., Ververas, E., Sun, G., Deng, J., Zafeiriou, S.: Neural sign actors: A diffusion model for 3d sign language production from text. arXiv preprint arXiv:2312.02702  (2023)

\bibitem{bruce2022mmnet}
Bruce, X., Liu, Y., Zhang, X., Zhong, S.h., Chan, K.C.: Mmnet: A model-based multimodal network for human action recognition in rgb-d videos. IEEE Transactions on Pattern Analysis and Machine Intelligence  \textbf{45}(3),  3522--3538 (2022)

\bibitem{Phoenix14t}
Camgoz, N.C., Hadfield, S., Koller, O., Ney, H., Bowden, R.: Neural sign language translation. In: Proceedings of the IEEE/CVF Conference on Computer Vision and Pattern Recognition (CVPR). pp. 7784--7793 (2018)

\bibitem{I3D}
Carreira, J., Zisserman, A.: Quo vadis, action recognition? a new model and the kinetics dataset. In: Proceedings of the IEEE/CVF Conference on Computer Vision and Pattern Recognition (CVPR). pp. 6299--6308 (2017)

\bibitem{chen2022two}
Chen, Y., Zuo, R., Wei, F., Wu, Y., Liu, S., Mak, B.: Two-stream network for sign language recognition and translation. In: Advances in Neural Information Processing Systems. vol.~35, pp. 17043--17056 (2022)

\bibitem{cox2002tessa}
Cox, S., Lincoln, M., Tryggvason, J., Nakisa, M., Wells, M., Tutt, M., Abbott, S.: Tessa, a system to aid communication with deaf people. In: Proceedings of the fifth international ACM conference on Assistive technologies. pp. 205--212 (2002)

\bibitem{cui2019deep}
Cui, R., Cao, Z., Pan, W., Zhang, C., Wang, J.: Deep gesture video generation with learning on regions of interest. IEEE Transactions on Multimedia  \textbf{22}(10),  2551--2563 (2019)

\bibitem{das2021vpn++}
Das, S., Dai, R., Yang, D., Bremond, F.: Vpn++: Rethinking video-pose embeddings for understanding activities of daily living. IEEE Transactions on Pattern Analysis and Machine Intelligence  \textbf{44}(12),  9703--9717 (2021)

\bibitem{das2020vpn}
Das, S., Sharma, S., Dai, R., Bremond, F., Thonnat, M.: Vpn: Learning video-pose embedding for activities of daily living. In: European Conference on Computer Vision (ECCV). pp. 72--90. Springer (2020)

\bibitem{fang2023signdiff}
Fang, S., Sui, C., Zhang, X., Tian, Y.: Signdiff: Learning diffusion models for american sign language production. arXiv preprint arXiv:2308.16082  (2023)

\bibitem{goodfellow2014generative}
Goodfellow, I., Pouget-Abadie, J., Mirza, M., Xu, B., Warde-Farley, D., Ozair, S., Courville, A., Bengio, Y.: Generative adversarial nets. In: Advances in Neural Information Processing Systems. vol.~27 (2014)

\bibitem{FID}
Heusel, M., Ramsauer, H., Unterthiner, T., Nessler, B., Hochreiter, S.: Gans trained by a two time-scale update rule converge to a local nash equilibrium. In: Proceedings of the 31st International Conference on Neural Information Processing Systems. pp. 6629--6640 (2017)

\bibitem{hu2023continuous}
Hu, L., Gao, L., Liu, Z., Feng, W.: Continuous sign language recognition with correlation network. In: Proceedings of the IEEE/CVF Conference on Computer Vision and Pattern Recognition (CVPR). pp. 2529--2539 (2023)

\bibitem{jia2022human}
Jia, Z., Zhang, Z., Wang, L., Tan, T.: Human image generation: A comprehensive survey. arXiv preprint arXiv:2212.08896  (2022)

\bibitem{jiang2021skeleton}
Jiang, S., Sun, B., Wang, L., Bai, Y., Li, K., Fu, Y.: Skeleton aware multi-modal sign language recognition. In: Proceedings of the IEEE/CVF Conference on Computer Vision and Pattern Recognition (CVPR). pp. 3413--3423 (2021)

\bibitem{HRNet}
Jin, S., Xu, L., Xu, J., Wang, C., Liu, W., Qian, C., Ouyang, W., Luo, P.: Whole-body human pose estimation in the wild. In: European Conference on Computer Vision (ECCV). pp. 196--214. Springer (2020)

\bibitem{kim2023cross}
Kim, S., Ahn, D., Ko, B.C.: Cross-modal learning with 3d deformable attention for action recognition. In: Proceedings of the IEEE/CVF International Conference on Computer Vision (ICCV). pp. 10265--10275 (2023)

\bibitem{kipp2011sign}
Kipp, M., Heloir, A., Nguyen, Q.: Sign language avatars: Animation and comprehensibility. In: Intelligent Virtual Agents: 10th International Conference, IVA 2011, Reykjavik, Iceland, September 15-17, 2011. Proceedings 11. pp. 113--126. Springer (2011)

\bibitem{kissel2021pose}
Kissel, C., K{\"u}mmel, C., Ritter, D., Hildebrand, K.: Pose-guided sign language video gan with dynamic lambda. arXiv preprint arXiv:2105.02742  (2021)

\bibitem{li2020word}
Li, D., Rodriguez, C., Yu, X., Li, H.: Word-level deep sign language recognition from video: A new large-scale dataset and methods comparison. In: The IEEE Winter Conference on Applications of Computer Vision. pp. 1459--1469 (2020)

\bibitem{li2020pona}
Li, K., Zhang, J., Liu, Y., Lai, Y.K., Dai, Q.: Pona: Pose-guided non-local attention for human pose transfer. IEEE Transactions on Image Processing  \textbf{29},  9584--9599 (2020)

\bibitem{li2019dense}
Li, Y., Huang, C., Loy, C.C.: Dense intrinsic appearance flow for human pose transfer. In: Proceedings of the IEEE/CVF Conference on Computer Vision and Pattern Recognition (CVPR). pp. 3693--3702 (2019)

\bibitem{liu2022dynast}
Liu, S., Ye, J., Ren, S., Wang, X.: Dynast: Dynamic sparse transformer for exemplar-guided image generation. In: European Conference on Computer Vision. pp. 72--90. Springer (2022)

\bibitem{liu2019liquid}
Liu, W., Piao, Z., Min, J., Luo, W., Ma, L., Gao, S.: Liquid warping gan: A unified framework for human motion imitation, appearance transfer and novel view synthesis. In: Proceedings of the IEEE/CVF International Conference on Computer Vision (ICCV). pp. 5904--5913 (2019)

\bibitem{ma2017pose}
Ma, L., Jia, X., Sun, Q., Schiele, B., Tuytelaars, T., Van~Gool, L.: Pose guided person image generation. In: Advances in neural information processing systems. vol.~30, pp. 405--415 (2017)

\bibitem{ma2021fda}
Ma, L., Huang, K., Wei, D., Ming, Z.Y., Shen, H.: Fda-gan: Flow-based dual attention gan for human pose transfer. {IEEE} Trans. Multimedia  \textbf{25},  930--941 (2023)

\bibitem{mcdonald2016automated}
McDonald, J., Wolfe, R., Schnepp, J., Hochgesang, J., Jamrozik, D.G., Stumbo, M., Berke, L., Bialek, M., Thomas, F.: An automated technique for real-time production of lifelike animations of american sign language. Universal Access in the Information Society  \textbf{15},  551--566 (2016)

\bibitem{ren2022neural}
Ren, Y., Fan, X., Li, G., Liu, S., Li, T.H.: Neural texture extraction and distribution for controllable person image synthesis. In: Proceedings of the IEEE/CVF Conference on Computer Vision and Pattern Recognition (CVPR). pp. 13535--13544 (2022)

\bibitem{ren2021combining}
Ren, Y., Wu, Y., Li, T.H., Liu, S., Li, G.: Combining attention with flow for person image synthesis. In: Proceedings of the 29th ACM International Conference on Multimedia. pp. 3737--3745 (2021)

\bibitem{ren2020deep}
Ren, Y., Yu, X., Chen, J., Li, T.H., Li, G.: Deep image spatial transformation for person image generation. In: Proceedings of the IEEE/CVF Conference on Computer Vision and Pattern Recognition (CVPR). pp. 7690--7699 (2020)

\bibitem{ronchetti2023lsa64}
Ronchetti, F., Quiroga, F., Estrebou, C., Lanzarini, L., Rosete, A.: Lsa64: A dataset of argentinian sign language. XX II Congreso Argentino de Ciencias de la Computación (CACIC)  (2016)

\bibitem{saunders2020everybody}
Saunders, B., Camgoz, N.C., Bowden, R.: Everybody sign now: Translating spoken language to photo realistic sign language video. arXiv preprint arXiv:2011.09846  (2020)

\bibitem{saunders2021continuous}
Saunders, B., Camgoz, N.C., Bowden, R.: Continuous 3d multi-channel sign language production via progressive transformers and mixture density networks. International Journal of Computer Vision  \textbf{129}(7),  2113--2135 (2021)

\bibitem{saunders2021mixed}
Saunders, B., Camgoz, N.C., Bowden, R.: Mixed signals: Sign language production via a mixture of motion primitives. In: Proceedings of the IEEE/CVF Conference on Computer Vision and Pattern Recognition (CVPR). pp. 1919--1929 (2021)

\bibitem{saunders2022signing}
Saunders, B., Camgoz, N.C., Bowden, R.: Signing at scale: Learning to co-articulate signs for large-scale photo-realistic sign language production. In: Proceedings of the IEEE/CVF Conference on Computer Vision and Pattern Recognition (CVPR). pp. 5141--5151 (2022)

\bibitem{saunders2021skeletal}
Saunders, B., Camg{\"o}z, N.C., Bowden, R.: Skeletal graph self-attention: Embedding a skeleton inductive bias into sign language production. In: Proceedings of the 7th International Workshop on Sign Language Translation and Avatar Technology: The Junction of the Visual and the Textual: Challenges and Perspectives. European Language Resources Association (Jun 2022)

\bibitem{siarohin2019animating}
Siarohin, A., Lathuili{\`e}re, S., Tulyakov, S., Ricci, E., Sebe, N.: Animating arbitrary objects via deep motion transfer. In: Proceedings of the IEEE/CVF Conference on Computer Vision and Pattern Recognition (CVPR). pp. 2377--2386 (2019)

\bibitem{siarohin2019first}
Siarohin, A., Lathuili{\`e}re, S., Tulyakov, S., Ricci, E., Sebe, N.: First order motion model for image animation. Advances in neural information processing systems  \textbf{32} (2019)

\bibitem{siarohin2021motion}
Siarohin, A., Woodford, O.J., Ren, J., Chai, M., Tulyakov, S.: Motion representations for articulated animation. In: Proceedings of the IEEE/CVF Conference on Computer Vision and Pattern Recognition (CVPR). pp. 13653--13662 (2021)

\bibitem{stoll2020text2sign}
Stoll, S., Camgoz, N.C., Hadfield, S., Bowden, R.: Text2sign: towards sign language production using neural machine translation and generative adversarial networks. International Journal of Computer Vision  \textbf{128}(4),  891--908 (2020)

\bibitem{Stoll_ACVR2020}
Stoll, S., Hadfield, S., Bowden, R.: Signsynth: Data-driven sign language video generation. In: European Conference on Computer Vision (ECCV). pp. 353--370. Springer (2020)

\bibitem{WER}
Su, K.Y., Wu, M.W., Chang, J.S.: A new quantitative quality measure for machine translation systems. In: COLING 1992 Volume 2: The 14th International Conference on Computational Linguistics (1992)

\bibitem{suo2022jointly}
Suo, Y., Zheng, Z., Wang, X., Zhang, B., Yang, Y.: Jointly harnessing prior structures and temporal consistency for sign language video generation. ACM Trans. Multimedia Comput. Commun. Appl.  (2024). \doi{10.1145/3648368}

\bibitem{L1}
Sural, S., Histogram, C., Distance, E., Distance, M., Cosine, V., Distance, A., Intersection, H., Ayyasamy, V., Majumdar, A.: Performance comparison of distance metrics in content-based image retrieval applications. Proc. of Internat. Conf. on Information Technology, Bhubaneswar, India pp. 159--164 (01 2003)

\bibitem{tang2020xinggan}
Tang, H., Bai, S., Zhang, L., Torr, P.H., Sebe, N.: Xinggan for person image generation. In: European Conference on Computer Vision (ECCV). pp. 717--734. Springer (2020)

\bibitem{tang2021structure}
Tang, J., Yuan, Y., Shao, T., Liu, Y., Wang, M., Zhou, K.: Structure-aware person image generation with pose decomposition and semantic correlation. In: Proceedings of the AAAI Conference on Artificial Intelligence. vol.~35, pp. 2656--2664 (2021)

\bibitem{tao2022structure}
Tao, J., Wang, B., Xu, B., Ge, T., Jiang, Y., Li, W., Duan, L.: Structure-aware motion transfer with deformable anchor model. In: Proceedings of the IEEE/CVF Conference on Computer Vision and Pattern Recognition (CVPR). pp. 3637--3646 (2022)

\bibitem{FVD}
Unterthiner, T., van Steenkiste, S., Kurach, K., Marinier, R., Michalski, M., Gelly, S.: Towards accurate generative models of video: A new metric \& challenges. arXiv preprint arXiv:1812.01717  (2019)

\bibitem{ventura2021everybody}
Ventura, L., Duarte, A., i~Nieto, X.G.: Can everybody sign now? exploring sign language video generation from 2d poses. arXiv preprint arXiv:2012.10941  (2021)

\bibitem{wang2022latent}
Wang, Y., Yang, D., Bremond, F., Dantcheva, A.: Latent image animator: Learning to animate images via latent space navigation. In: International Conference on Learning Representations (ICLR) (2022)

\bibitem{SSIM}
Wang, Z., Bovik, A., Sheikh, H., Simoncelli, E.: Image quality assessment: from error visibility to structural similarity. IEEE Transactions on Image Processing  \textbf{13}(4),  600--612 (2004)

\bibitem{zhang2022exploring}
Zhang, P., Yang, L., Lai, J.H., Xie, X.: Exploring dual-task correlation for pose guided person image generation. In: Proceedings of the IEEE/CVF Conference on Computer Vision and Pattern Recognition (CVPR). pp. 7713--7722 (2022)

\bibitem{LPIPS}
Zhang, R., Isola, P., Efros, A.A., Shechtman, E., Wang, O.: The unreasonable effectiveness of deep features as a perceptual metric. In: Proceedings of the IEEE/CVF Conference on Computer Vision and Pattern Recognition (CVPR). pp. 586--595 (2018)

\bibitem{zhao2022thin}
Zhao, J., Zhang, H.: Thin-plate spline motion model for image animation. In: Proceedings of the IEEE/CVF Conference on Computer Vision and Pattern Recognition (CVPR). pp. 3657--3666 (2022)

\bibitem{zhou2021improving}
Zhou, H., Zhou, W., Qi, W., Pu, J., Li, H.: Improving sign language translation with monolingual data by sign back-translation. In: Proceedings of the IEEE/CVF Conference on Computer Vision and Pattern Recognition (CVPR). pp. 1316--1325 (2021)

\bibitem{zuo2024simple}
Zuo, R., Wei, F., Chen, Z., Mak, B., Yang, J., Tong, X.: A simple baseline for spoken language to sign language translation with 3d avatars. arXiv preprint arXiv:2401.04730  (2024)

\end{thebibliography}

\end{document}